\documentclass[runningheads]{llncs}
\usepackage[T1]{fontenc}
\usepackage{amsmath}
\usepackage{amssymb}
\usepackage{subcaption}
\usepackage{xcolor}
\usepackage{graphicx}
\usepackage[ruled, vlined, linesnumbered]{algorithm2e}
\begin{document}
\title{Offline Deep Model Predictive Control (MPC) for Visual Navigation}
% \author{Taha Bouzid \inst{1} \and Youssef Alj\inst{1}}
\author{Taha Bouzid\inst{1,2}\and
Youssef Alj\inst{1}}
\institute{International Artificial Intelligence Center of Morocco, Ai movement\\Mohammed VI Polytechnic University, Rabat, Morocco\and
ENSAM, Moulay Ismail University, Meknes, Morocco}

%email{lncs@springer.com}\\
\maketitle              % typeset the header of the contribution
\begin{abstract}
In this paper, we propose a new visual navigation method based on a single RGB perspective camera. Using the Visual Teach \& Repeat (VT\&R) methodology \cite{furgale2010visual}, the robot acquires a visual trajectory consisting of multiple subgoal images in the teaching step. In the repeat step, we propose two network architectures, namely ViewNet and VelocityNet. The combination of the two networks allows the robot to follow the visual trajectory. ViewNet is trained to generate a future image based on the current view and the velocity command. The generated future image is combined with the subgoal image for training VelocityNet. We develop an offline Model Predictive Control (MPC) policy within VelocityNet with the dual goals of (1) reducing the difference between current and subgoal images and (2) ensuring smooth trajectories by mitigating velocity discontinuities. Offline training conserves computational resources, making it a more suitable option for scenarios with limited computational capabilities, such as embedded systems. We validate our experiments in a simulation environment, demonstrating that our model can effectively minimize the metric error between real and played trajectories.

\end{abstract}

\section{INTRODUCTION}
Visual navigation is the task of guiding a robot toward a specific goal using an RGB camera as its primary exteroceptive sensor. In order to achieve this task, existing techniques can be categorized into two main groups: map-based navigation and reactive methods. In the former, the robot relies on a pre-built map representing the environment and tries to reach the goal. In the latter, the robot makes decisions, such as obstacle avoidance, on the fly, without prior knowledge of the environment. In the context of map-based visual navigation, several challenges require careful consideration. First, map-building: The construction of an accurate and informative map is a fundamental step in map-based navigation. These maps can take various forms, including topological graphs \cite{kwon2021visual}, metric maps derived for example from SLAM-based techniques \cite{chung2023orbeez}, or more recent neural implicit representations \cite{zhu2022nice}. Secondly, localization: Precise self-localization within the map is essential for effective navigation. To ensure that the robot knows its position accurately within the environment, localization methods must seamlessly integrate with the chosen map representation. Thirdly, path planning: Efficient global and local path planning strategies are essential for guiding the robot from its current location to the desired goal. These strategies must navigate through the map complexity while optimizing some criteria such as distance, safety, and energy efficiency. Finally, control algorithms: Control algorithms play a crucial role in executing the planned path. These algorithms generate velocities and control signals that are sent to the robot actuators, translating planned actions into physical movement. While control algorithms are critical, they often operate within the context of the chosen navigation approach.

One widely adopted approach for robot control is visual servoing \cite{chaumette2006visual}, in which the primary objective is to guide the robot toward a desired image using continuous visual feedback. However, visual servoing alone may not inherently address obstacle avoidance, necessitating dedicated routines. Additionally, the non-linear nature of the cost function within the visual servoing framework can limit the convergence domain. In recent years, learning-based methods \cite{zhu2017target,hirose2019deep,gupta2017cognitive} have gained an increasing interest. Reinforcement Learning (RL) \cite{zhu2017target}, for example, offers a framework where an agent is trained to learn a policy that maps visual observations directly to actions. RL presents substantial challenges, often requiring an extensive number of trial-and-error experiments to converge successfully. While initial training can be conducted in a simulated environment, adapting these learned policies to real-world scenarios remains a complex and ongoing challenge \cite{zhao2020sim}. In contrast, Model Predictive Control (MPC), enables a robot to navigate through its environment by continuously synthesizing future images based on its current state and then planning control actions to achieve desired visual goals. MPC has found applications in diverse problem domains, including trajectory planning \cite{finn2017deep}, humanoid gait generation \cite{tanguy2019closed}, and wheeled robot control \cite{hirose2019deep}. 

The main contributions of our work are twofold. First, it introduces VelocityNet as an offline deep MPC navigation system. Second, we adapt the existing network architectures to a single perspective camera. Offline training offers several noteworthy benefits, including enhanced safety, cost-effectiveness, and greater adaptability. It allows training within a simulated environment, which is safer than training the robot in the real world. The approach also has the advantage of being time-efficient. This is because the robot only needs to collect data from its environment rather than undergo training within it, which significantly reduces the time required for the overall process. Offline training does not require access to the real robot, which can save money on hardware and setup costs. We will demonstrate our model's ability to generate precise velocities for a robot to follow a desired trajectory with minimal errors in various scenarios.

This paper is organized as follows: In section~\ref{sec:related_work}, we first present work related to visual navigation. Section~\ref{sec:proposed_approach} begins with a description of the simulation environment, followed by an explanation of the architecture and training process of the two proposed network architectures, namely, ViewNet and VelocityNet. In Section~\ref{sec:exp_setup}, we define the experimental setup employed in our study. Moving on to section~\ref{sec:results_analysis}, we present the results of our simulations under various motion types. Finally, in Section~\ref{sec:conclusion}, we wrap up the paper with our concluding remarks.

\section{RELATED WORK}
\label{sec:related_work}
In this section, we introduce three key techniques for visual navigation: Visual Servoing, Reinforcement Learning (RL), and Model Predictive Control (MPC).

\paragraph{\textbf{Visual Servoing:}}
Visual servoing (VS) methods have a long history in robotics~\cite{chaumette2006visual}, VS is a closed-loop control technique that uses real-time visual feedback, primarily in the form of extracted image features, to guide the motion or position of a robot or camera. This technique allows a wide variety of tasks designed to locate a system with respect to its environment or to reach a target image, by controlling one or several degrees of freedom. Optimization methods are employed in order to find the optimal control inputs that minimize the error between the desired visual features and the observed ones. VS has been used in the visual path following \cite{diosi2007outdoor,diosi2011experimental}. A visual path is represented by a series of target images and a visual servoing algorithm allows the robot to navigate from its current position to the next key image. VS suffers from a limited convergence area due to a highly non-linear cost function. Furthermore, it is very sensitive to lighting conditions and requires accurate camera calibration. With the development of deep learning architectures, VS can be used to control a robot in a latent space as proposed in \cite{felton2023deep}, where the representations of camera pose and the embeddings of their respective images are tied together. This allows a larger convergence domain and more accurate robot positioning.

\paragraph{\textbf{Reinforcement Learning (RL) for Visual Navigation:}}

In recent years, the rise of Reinforcement Learning (RL) has revolutionized the field of robotics, particularly in the context of visual navigation. RL has demonstrated its capability to learn control policies directly from visual observations \cite{zhu2017target}, offering the potential to navigate to distant goals while optimizing user-specified reward functions. 
Researchers have investigated RL-based approaches, training agents to navigate through environments using simulated experiences. 
Online reinforcement learning for robotic navigation is difficult to implement in real-world settings, as it requires several experiments of trial and error. Offline reinforcement learning methods, which can leverage existing datasets of robotic navigation data, offer a more scalable and generalizable approach. Shah et al. \cite{shah2022offline} proposed an offline reinforcement learning system for robotic navigation that can optimize user-specified reward functions in the real world, their use of graph search in combination with RL parallels prior work that integrates planning into supervised skill learning methods \cite{shah2022offline,savinov2018semiparametric} and goal-conditioned reinforcement learning \cite{NEURIPS2019_5c48ff18}. The transition from simulation to the real world remains a challenging task and requires extensive domain adaptation and robustness to unforeseen real-world conditions. Nevertheless, RL in visual navigation is a promising approach, and the field continues to evolve with innovations in model architectures, training methodologies, and transfer learning techniques.
Our work differs from those previous approaches as it focuses on a subset of visual navigation problems known as visual path following. In this approach, there is no need for trajectory planning or graph optimization. Instead, the robot is provided with a trajectory composed of images, referred to as subgoals, which it is expected to follow.
\paragraph{\textbf{Model Predictive Control (MPC) in Visual Navigation:}}

Model Predictive Control has emerged as a powerful paradigm in visual navigation, allowing robots to predict future images based on their current state and plan control actions accordingly. This approach transcends traditional control strategies, enabling robots to anticipate changes in the environment and adapt their actions proactively. MPC-based methods find applications in various facets of visual navigation, such as trajectory planning \cite{10.1145/3478586.3478601}, obstacle avoidance, and path following \cite{YU2018247}. By explicitly accounting for visual feedback and uncertainty, MPC enhances an agent's capability to navigate dynamically in changing environments with agility and precision. Previous approach \cite{hirose2019deep} used an Online Deep Visual MPC-policy learning method that can perform visual navigation while avoiding collisions with unseen objects on the navigation path. However such Online learning methods require a large amount of online experience. The main goal of our approach is to develop an offline Deep Visual MPC-policy learning method and to explore its application to robotic navigation tasks by building a learning-based system for long-horizon planning.

%%%%%%%%%%%%%%%%%%%%%%%%%%%%%%%%%%%%%%%%%%%%%%%%%%%%%%%%%
\section{PROPOSED APPROACH}
\label{sec:proposed_approach}
%%%%%%%%%%%%%%%%%%%%%%%%%%%%%%%%%%%%%%%%%%%%%%%%%%%%
In this section, we will present the simulated environment and the network architectures used to control the robot, as well as the training process of these networks.
\subsection{Simulation environment}
The simulation was conducted in ROS Gazebo. The simulated environment comprises two key models: a house and a robot. Within this environment, the robot is capable of navigating inside the house and collecting the necessary images for training our models.
\subsubsection{The house model:}
\begin{figure}[!h]
    \centering
    \includegraphics[width=\linewidth]{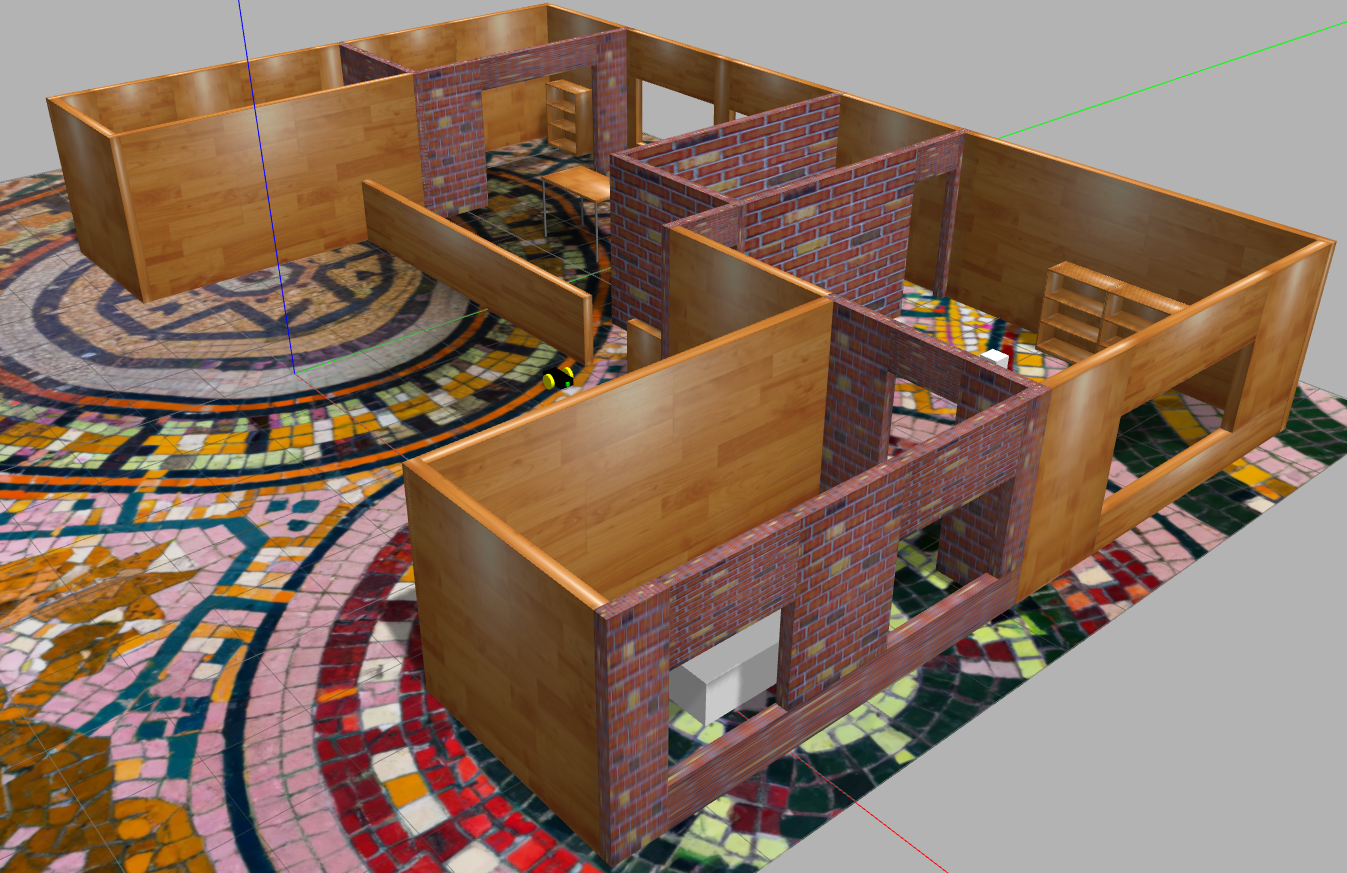}
    \caption{Upper view of the house model}
    \label{House}
\end{figure}

We use the house represented in Figure~\ref{House} for the robot navigation. The robot can move inside this model, capture images, and receive velocity commands to collect a dataset that will be used for training. The house is composed of different rooms, each containing various items of furniture. These items contribute to the photometric richness of the captured images.

We will use a floor that features the shape and colors of a mosaic as shown in Figure~\ref{House}. In this manner, the floor will contain various colors and shapes, which will significantly enhance the richness of our model. It's important to note that our models employ CNN layers for feature extraction. Consequently, providing them with a multitude of features can augment their capabilities.
\begin{figure}[!h]
  \begin{subfigure}{0.45\textwidth}  % First subfigure
    \centering
    \includegraphics[width=\linewidth]{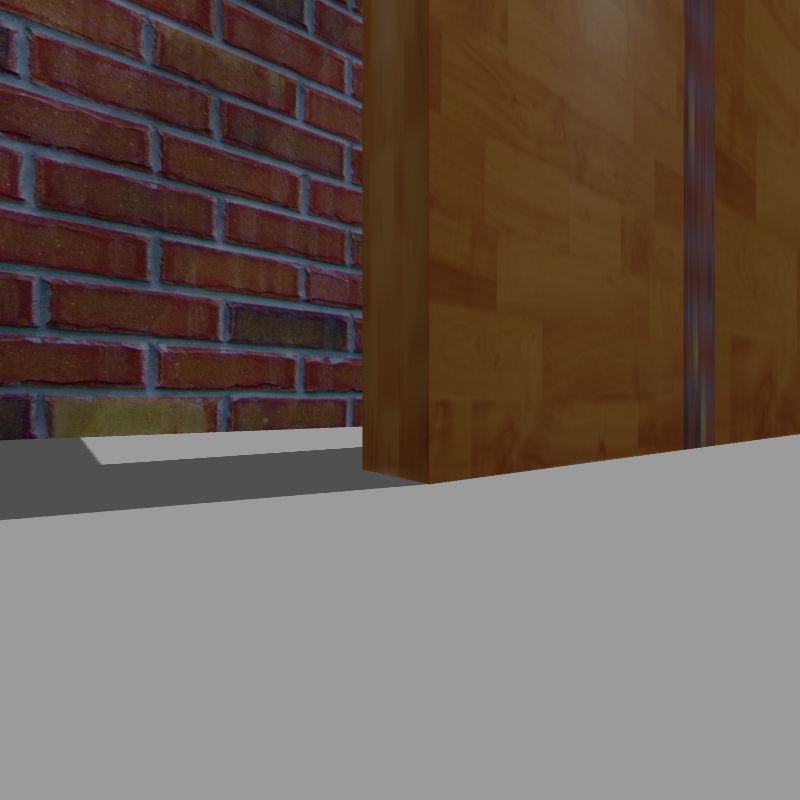}
    \caption{}
    \label{fig:left_image}
  \end{subfigure}%
  \quad
  \begin{subfigure}{0.45\textwidth}  % Second subfigure
    \centering
    \includegraphics[width=\linewidth]{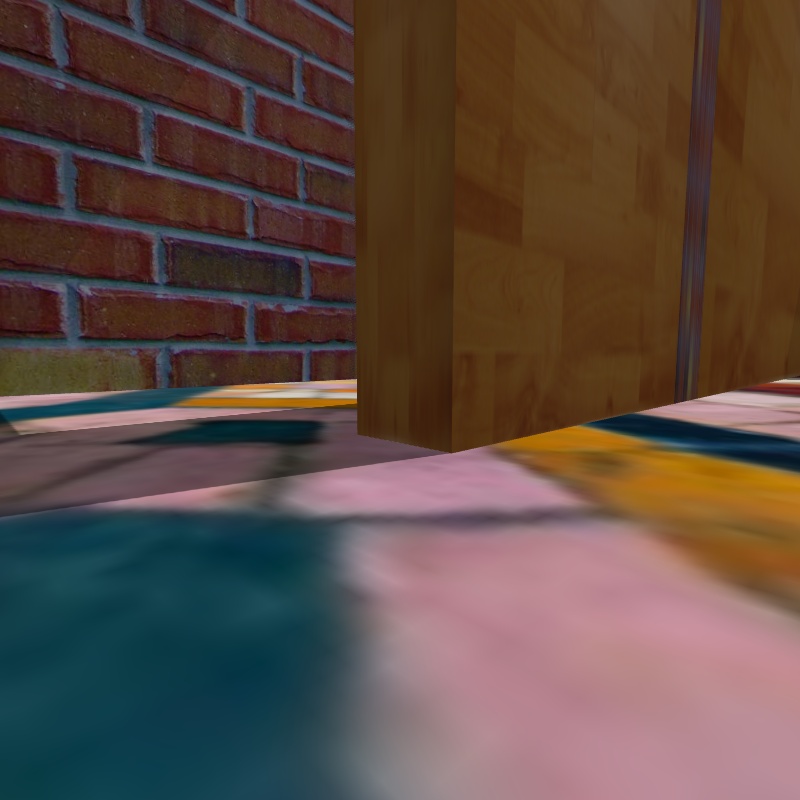}
    \caption{}
    \label{fig:right_image}
  \end{subfigure}
  \caption{Images captured from the robot perspective camera, depicting 
  the changes made to the scene by texturing the floor, (a) before texturing, and (b) after texturing.}
  \label{fig:both_images}
\end{figure}

Figure~\ref{fig:both_images} shows the difference between the floor with and without changes, the right image depicts an example of the images that will be used for the training of our model.

\subsubsection{The robot model:}
\begin{figure}[!h]
      \centering
      \includegraphics[width=0.5\linewidth]{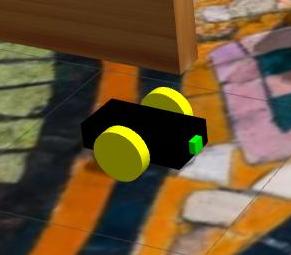}
      \caption{The mobile robot model used in the simulation}
      \label{fig:robot}
\end{figure}

The mobile robot model represented in Figure~\ref{fig:robot} that we use in our simulation can move in a 2D space with only two velocities: longitudinal velocity denoted as $v_x$ and angular velocity around the z-axis denoted as $\omega_z$. From now on, for the sake of simplicity, we denote $v_x$ and $\omega_z$ as $v$ and $\omega$ respectively; We can move our robot by sending twist velocity messages to the corresponding topic dedicated to moving the robot. We attach a forward-looking camera in the front of the robot enabling it to take images from the environment. Hence, we can acquire the dataset images by simulating robot movement within the environment. To extract velocity commands, we can access the twist messages published in the corresponding velocity topic and subsequently link each image to its corresponding velocity command.

\subsection{Future image prediction, ViewNet }
\begin{figure}[!h]
  \centering
  \includegraphics[width=\linewidth]{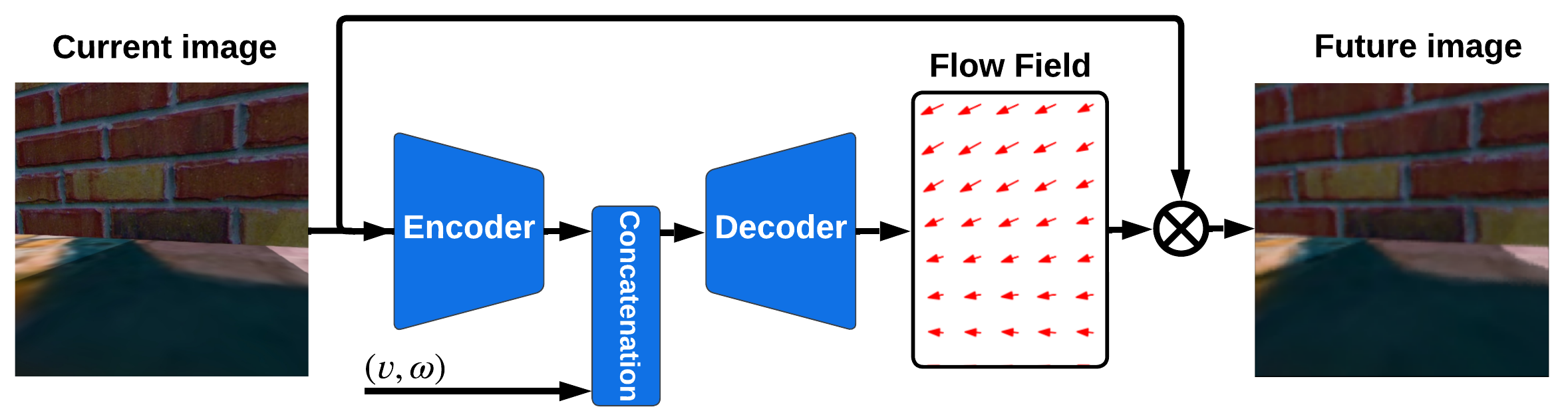}
  \caption{ViewNet architecture for future image prediction.}
  \label{encoder}
\end{figure}

Figure~\ref{encoder} represents ViewNet architecture inspired from \cite{hirose2019vunet}. The network takes as input an image captured from our camera's perspective and a given velocity $\xi_t=(v, \omega)$ and outputs a prediction image. The architecture is based on the Encoder-Decoder (ED) architecture depicted in Figure~\ref{encoder}. The ED takes two inputs: the image input is directly sent to the encoder and the velocity is concatenated with the output of the encoder.
The output of the decoder is a two-dimensional flow field image as represented in Figure~\ref{encoder}. The flow field is used to sample the input image and generate the future prediction image. Images synthesized with ViewNet will be used to train VelocityNet within the MPC policy defined in Equations~\eqref{subsec:img_loss1} and~\eqref{subsec:img_loss2}.

\subsection{Control Policy with VelocityNet}
We present a novel navigation system based only on perspective images from an RGB camera. Our model can generate velocities necessary for a mobile robot to reach subgoal images. 

\begin{figure}[!h]
    \centering
    \includegraphics[width=\linewidth]{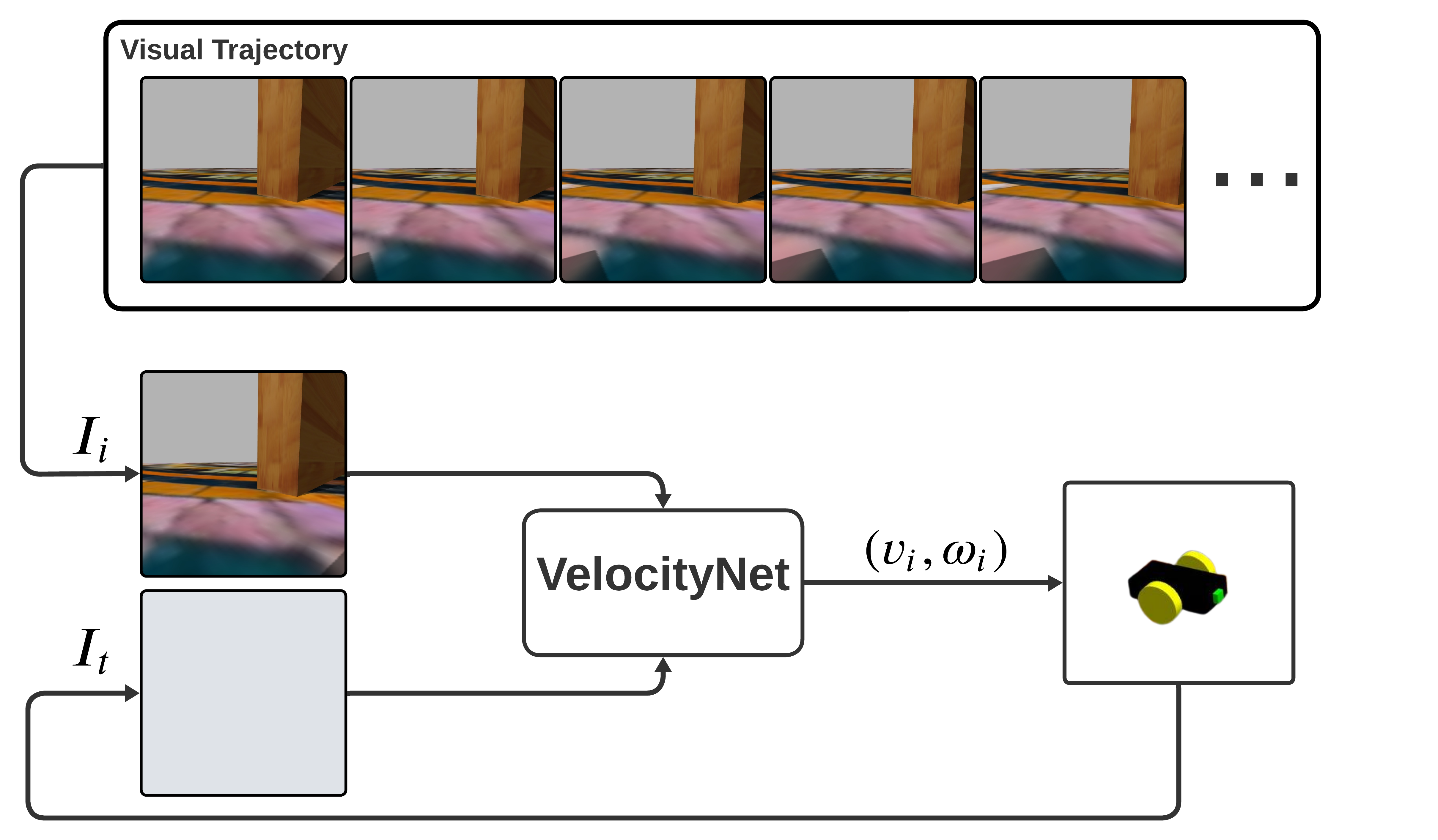}
    \caption{The proposed path following approach. $I_t$ is the image from the camera attached to the front of the robot at time $t$, $I_i$ is the subgoal image of index $i$ from the visual trajectory, $(v_i,\omega_i)$ are the generated velocities commands to reach the $i$-th subgoal image.}
    \label{fig:path_following}
\end{figure}

Figure~\ref{fig:path_following} shows the proposed architecture for path following. At time $t$, VelocityNet takes the robot current image from the camera and the subgoal image as inputs. The visual trajectory is defined by $l$ subgoal images $(I_0, I_1, ..., I_{l-1})$ collected during a previous step, also referred to as the visual teach step. 

In the repeat step, the main objective of VelocityNet is to minimize the difference between the current image and the subgoal image at time $t$. By generating accurate velocities, the robot moves towards the location of the subgoal causing the robot's current camera image to resemble the subgoal image.

The robot can navigate along a specific trajectory by sequentially following subgoal images. When the condition $\left|I_{t}-I_{i}\right| \leq e_{m}$ is met the model assigns the next subgoal with index $i+1$, This condition indicates that the absolute image difference between the current image $I_t$ and the $i$-th subgoal image $I_i$ is smaller than the threshold $e_m$. We can experimentally choose a certain value for $e_m$ based on the expected precision and the time required for the robot to follow the entire trajectory. This method allows our robot to follow any given trajectory relying only on visual data from the RGB camera.

Typically, a topological graph is constructed, and the nearest image to the current one is selected from the visual trajectory to initiate the navigation. In our scenario, the robot starts visual navigation from an initial position corresponding to the first image, denoted as $I_0$, in the visual trajectory. However, the determination of whether the robot should move or not in the presence of a significant disparity between the real-time image and the subgoal image $I_0$ is governed by Algorithm~\ref{alg:follow-visual-trajectory}. This algorithm decides whether the robot should navigate to the initial image or proceed to the subsequent subgoal images.

\subsection{Training VelocityNet}

\begin{figure}[h!]
    \centering
    \includegraphics[width=0.7\linewidth]{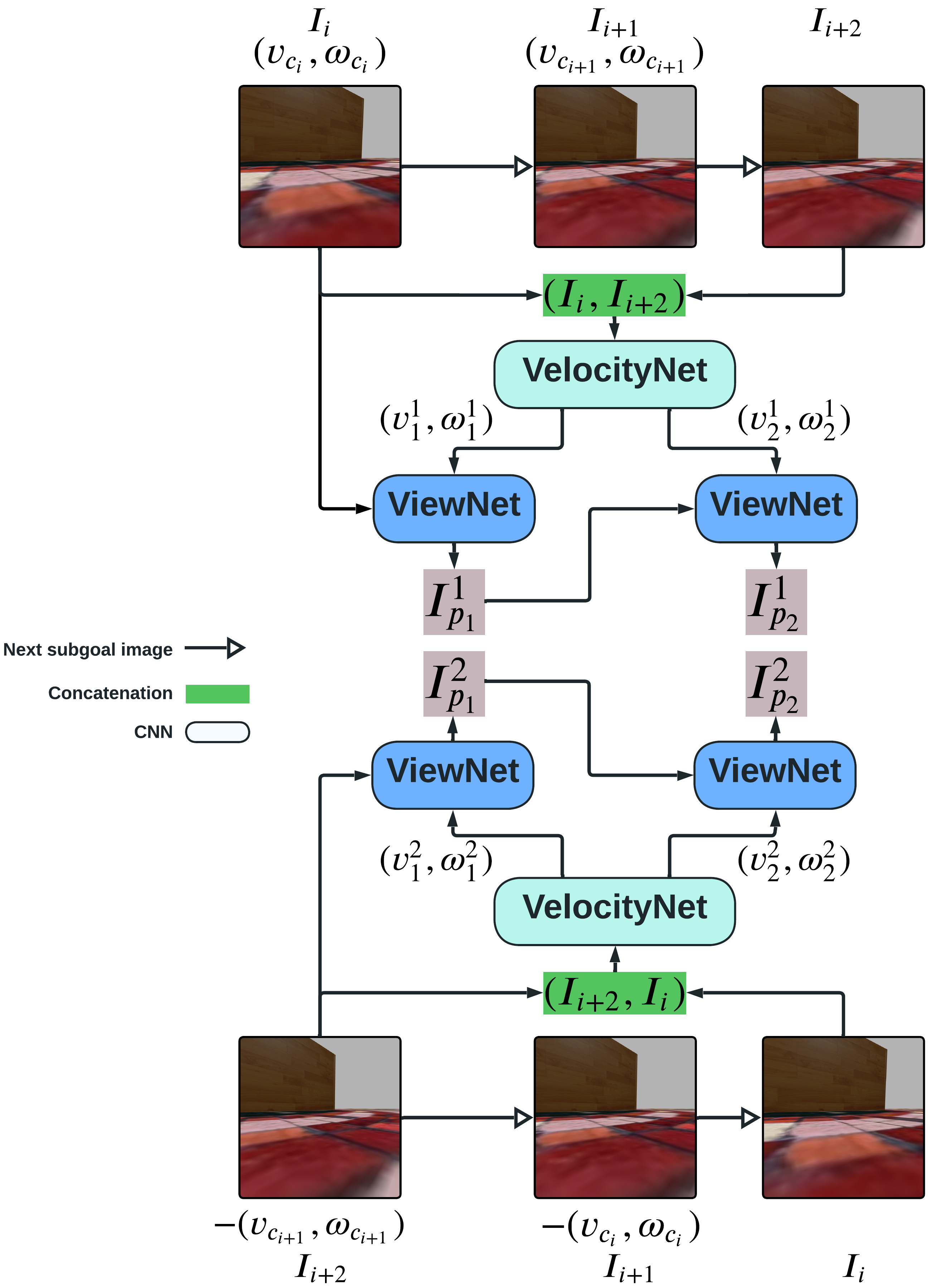}
    \caption{Offline training process of VelocityNet, the generated velocities and images are used for the loss calculation to update the model.}
    \label{fig:offline_training}
\end{figure}

We employ an offline deep Model Predictive Control (MPC) Policy to train VelocityNet, leveraging existing data collected for training ViewNet. This dataset includes images paired with corresponding velocity commands. As shown in Figure~\ref{fig:offline_training}, our training process involves using three consecutive images from the dataset $(I_{i}, I_{i+1}, I_{i+2})$ and with their corresponding velocity commands, denoted as $(v_{c_i}, \omega_{c_i})$ and $(v_{c_i+1}, \omega_{c_i+1})$, associated with the initial and second images. The index $i$ can be chosen randomly within the range of subgoal indexes. These velocity commands serve as the control inputs that guide the robot movement from the current image, denoted as $I_i$, towards the subgoal image $I_{i+2}$ during data collection.

VelocityNet undergoes two forward calculations during training, resulting in two sets of velocities: $(v_1^{j}, \omega_1^{j})$ and $(v_2^{j}, \omega_2^{j})$, distinguished by the index 'j' ('1' for the upper one and '2' for the other). During the initial forward calculation, we concatenate $I_i$ with $I_{i+2}$ and feed this concatenated data into VelocityNet. The order of concatenation is significant: the first element represents the current image, and the second one depicts the goal image.

In the subsequent forward calculation, we reverse the roles of consecutive images $I_{i}$, $I_{i+1}$, and $I_{i+2}$, along with their corresponding velocity commands, such that $I_{i+2}$ becomes the current image and $I_{i}$ becomes the subgoal image. Consequently, the corresponding velocity commands become $-(v_{c_{i+1}},\omega_{c_{i+1}})$ and $-(v_{c_{i}},\omega_{c_{i}})$.

In fact, this second forward calculation was added to the model as a form of data augmentation, aiming to enhance the model's adaptability to various scenarios when dealing with different subgoals. This augmentation helps the model predict accurate velocities. 

The generated velocities are used in predicting images via ViewNet, which takes both velocities and images as input, producing prediction images. 

The predicted images $(I_{p_1}^{j}, I_{p_2}^{j})$ and the subgoal images $(I_{i}, I_{i+1}, I_{i+2})$ are used to compute the image loss as defined in Equations~\eqref{subsec:img_loss1} and~\eqref{subsec:img_loss2}. On the other hand, the generated velocities $((v_1^{j}, \omega_1^{j})$, $(v_2^{j}, \omega_2^{j}))$ and the velocity commands are employed to calculate the velocity loss as represented in Equations~\eqref{subsec:velo_loss1} and~\eqref{subsec:velo_loss2}.

 We define the cost function that corresponds to each calculation as a weighted combination of two components, the image loss $J_{img_{j}}$ and the velocity loss $J_{vel_j}$. This function penalizes large velocity changes, which can lead to jerky motion or instability. The defined MPC objective uses a horizon of $N$ time steps. The calculation of the best $N$ upcoming velocity commands involves minimizing the following cost:
\begin{equation}
J_j = w_1 J_{img_{j}}  + w_2 J_{vel_j}
\end{equation}   

where $w_1$ and $w_2$ are constant weights and $j$ the index of the forward calculation. The values of $w_1$ and $w_2$ can be adjusted to find the right balance between the two objectives. For example, if we want to prioritize smoothness, we can increase the value of $w_2$. If we want to prioritize accuracy, we can increase the value of $w_1$.
Nonetheless, a larger value of $w_2$ can make the model memorize the velocity commands linked to each image. However, the intention behind this supplementary factor is not to instruct the model to mimic the precise velocity commands of the robot but to regularize and obtain smooth velocities. 
\subsubsection{Image loss:} The image loss can be defined as the mean of absolute pixel difference between the subgoal images  ($I_{i}$, $I_{i+1}$, $I_{i+2}$)  and the sequence of predicted images $(I_{p_1}^{j},I_{p_2}^{j})$, the image loss for $j=1$ can be defined as: 
\begin{equation}
J_{img_{1}} =\frac{1}{2 n_{p i x}} \left(\left|I_{i+1}-I_{p_1}^{1}\right|+\left|I_{i+2}-I_{p_2}^{1}\right|\right)
\label{subsec:img_loss1}
\end{equation}
where $|.|$ denotes the Mean Absolute Error (MAE) function. This function outputs a single value representing the average absolute difference between predicted and subgoal images.
Image loss for $j=2$ can be defined as:
\begin{equation}
J_{img_{2}} =\frac{1}{2 n_{p i x}} \left(\left|I_{i+1}-I_{p_1}^{2}\right|+\left|I_{i}-I_{p_2}^{2}\right|\right)
\label{subsec:img_loss2}
\end{equation}

with $(I_{p_1}^{j},I_{p_2}^{j})$ are the predicted images generated by the model conditioned on virtual velocities, and $n_{pix}$ represents the number of pixels in the image, $3\times128\times128$.

\subsubsection{Velocity loss:}
\label{subsec:velo_loss}
Training our model based on the image loss alone can help our robot navigate precisely in the environment, but the velocities generated in that case can be fluctuating and non-realistic. As a solution, we propose to add a velocity loss that can help our model generate smooth velocities and minimize the difference between the generated velocities $((v_{1}^{j},w_{1}^{j}),(v_{2}^{j},w_{2}^{j}))$ and the velocity commands $((v_{c_i},\omega_{c_i})$, $(v_{c_{i+1}},\omega_{c_{i+1}}))$, the Velocity loss for $j=1$ can be defined as:
\begin{equation}
J_{vel_1} = \frac{1}{N} \sum_{n=1}^{N} \left((v_{c_{i-1+n}} - v_{n}^1)^2 +  (\omega_{c_{i-1+n}} - \omega_{n}^1)^2\right)
\label{subsec:velo_loss1}
\end{equation}

Velocity loss for $j=2$ can be defined as:
\begin{equation}
J_{vel_2} = \frac{1}{N} \sum_{n=1}^{N} \left((v_{c_{i+2-n}} + v_{n}^2)^2 + (\omega_{c_{i+2-n}} + \omega_{n}^2)^2\right)
\label{subsec:velo_loss2}
\end{equation}
The loss functions described in Equations~\eqref{subsec:velo_loss1} and~\eqref{subsec:velo_loss2} employ a normalized summation approach, combining terms with different units. Typically, a weighted sum is employed to prevent one term from overshadowing another due to differences in scale. However, in our formulation, we deliberately opted for equal weighting between the two terms with distinct units. This choice signifies that irrespective of their units, each term contributes equally to the overall loss. The absence of a dominant unit indicates that the chosen formulation successfully mitigates any disproportionate influence of one physical quantity over another.

We define the loss function that will be used in backpropagation as: 
\begin{equation}
J =\frac{1}{2} \sum_{j=1}^{N} J_j 
\end{equation} 

\subsection{Visual Trajectory Following with VelocityNet, Control Algorithm}

In Algorithm~\ref{alg:follow-visual-trajectory}, we describe the control algorithm used to follow the visual trajectory. During real-time operation, the mobile robot will use only the VelocityNet model to navigate between $l$ subgoals. Our model generates two sets of velocities, $(v_1, w_1)$ and $(v_2, w_2)$. The velocity command given to the robot is the first one generated by the model, $(v_1, w_1)$. If the loss condition, $\lvert I_{t} - I_{i} \rvert > e_{m}$ is satisfied, we wait for the robot to execute the twist message for a duration of 0.1 seconds, and then we send a null-velocity command to initiate a pause while waiting for the VelocityNet model to be calculated. Otherwise, we will switch to the next subgoal and the robot remains stationary, as we've previously published a null velocity command. Thereby we continue this process until we complete the visual trajectory.

\begin{algorithm}[!h]
\caption{Visual path following algorithm}
\label{alg:follow-visual-trajectory}
\KwData{
  Visual trajectory: A sequence of desired visual states or images $(I_{i})$\;
  Real-time image: The current image or visual state captured by the robot's camera $I_{t}$\;
}
\KwResult{Control the robot's motion to follow the visual trajectory}
 initialization\;
\While{subgoal index $<$ $l$}{
  calculate the l1 error between the subgoal and current image\;
  \eIf{$\lvert I_{t} - I_{i} \rvert > e_{m}$}{
    $v,\omega$ = VelocityNet(Real-time image, Subgoal image)\;
    publish($v,\omega$)\;
    moving the robot for 0.1 seconds\;
    $v$ = 0.0\;
    $\omega$ = 0.0\;
    publish($v,\omega$)\;
    
  }
  {
    subgoal index += 1\;
  }
}
\end{algorithm}

%%%%%%%%%%%%%%%%%%%%%%%%%%%%%%%%%%%%%%%%%%%
\section{EXPERIMENTAL SETUP}
\label{sec:exp_setup}
%%%%%%%%%%%%%%%%%%%%%%%%%%%%%%%%%%%%%%%%%%%%%
\subsection{Network Structure}
\subsubsection{ViewNet:}

ViewNet architecture is based on the Encoder-Decoder (ED) architecture. The encoder is constructed by 8 convolutional layers with batch normalization and leaky ReLU function. The encoder produces a 512-dimensional feature vector, which we combine with the two-dimensional velocity vector $(v,\omega)$ before inputting it into the decoder. Subsequently, the decoder generates a flow field image with dimensions $2 \times 128 \times 128$ that we use for sampling the input image. The output image from ViewNet is a three-channel (RGB) $128 \times 128$ image.
\subsubsection{VelocityNet:}
VelocityNet can generate $N$ steps robot velocities. Concatenated real-time image and subgoal image are input to 8 convolutional layers with batch-normalization and leaky ReLU activation functions, excluding the last layer. In the final layer, the feature is split into two $(N \times 1)$ vectors.
 The $\mathrm{tanh}(.)$ function is applied to the last layer of VelocityNet to keep the linear velocity within the range of $\pm v_{\text{max}}$ and the angular velocity within the range of $\pm \omega_{\text{max}}$, where $v_{\text{max}} = 0.5 \, \mathrm{m/s}$ and $\omega_{\text{max}} = 1.0 \, \mathrm{rad/s}$.
 
\subsection{Training}
We use the simulator previously described to train and evaluate our model based on our MPC approach.
VelocityNet and ViewNet are trained on the data collected from the simulation environment by moving the robot in the house model. We collect a total of 7024 images and the corresponding velocity commands. The collected images are of size $800 \times 800$, which are then resized to $128 \times 128$ before being fed into the network.

The training was performed on a computing infrastructure equipped with NVIDIA T4 GPUs. We train the networks separately, starting with ViewNet. They are both trained using the Adam optimizer with a learning rate of $1e-4$.
\subsection{Parameters}
We fix the horizon period to $N=2$ to limit error accumulation. As shown in Figure~\ref{fig:offline_training}, ViewNet takes input images from the dataset or predicted images generated from the previous ViewNet output. Therefore, using ViewNet for a long horizon will cause errors that may accumulate and result in predicted images that differ from the desired ones.

We set the weights $w_1$ and $w_2$ in the cost function $J_j$ to 0.8 and 0.3, respectively. We gave the weights those values because the robot is not allowed to memorize the exact velocity commands, but rather to smooth out the velocities generated by the model while prioritizing tracking subgoal images precisely.

The metric error $e_m$ for the subgoal switching condition, $\lvert I_{t} - I_{i} \rvert < e_{m}$, is set empirically to 17. This value represents an optimal choice for achieving precise tracking of subgoals and minimizing the time required to complete the visual trajectory.

Our robot will use only real-time images from the image topic and does not require any additional information from the environment. 

%%%%%%%%%%%%%%%%%%%%%%%%%%%%%%%%%%%%%%%%%%%%%%%%%%%%
\section{RESULTS ANALYSIS}
\label{sec:results_analysis}
%%%%%%%%%%%%%%%%%%%%%%%%%%%%%%%%%%%%%%%%%%%%%%%%%%%%
Model testing is performed on different scenarios. We collect three different visual trajectories: the first trajectory represents images corresponding to pure translation, the second one corresponds to pure rotation, and the third combines rotation and translation.

We evaluate our model based on these three scenarios and demonstrate that it can perform its required tasks with minimal errors. 

\subsection{Linear translation} 
\begin{figure}[!h]
    \centering
    \includegraphics[width=0.8\linewidth]{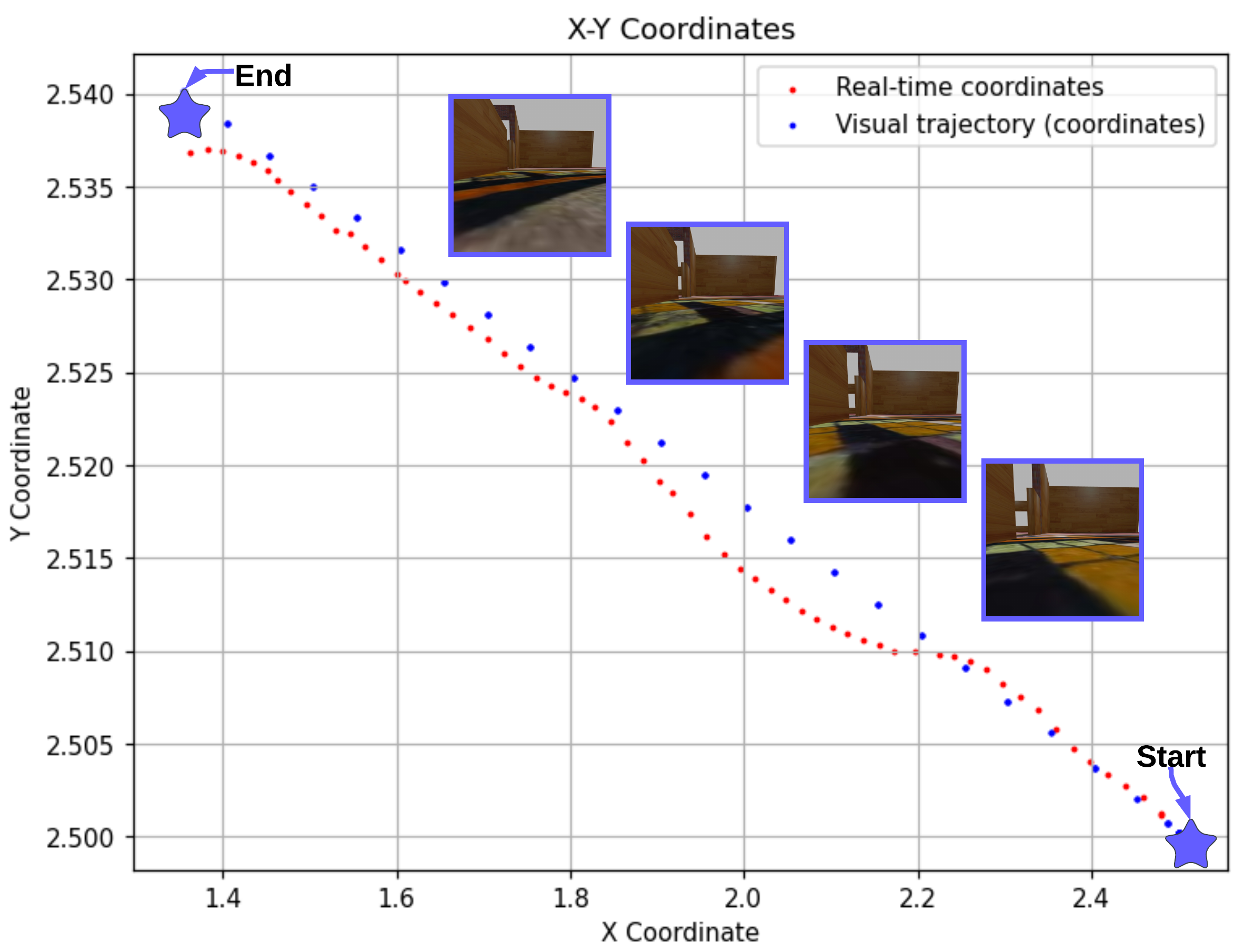}\caption{Evaluation of our predictive model VelocityNet in a linear translation scenario, blue points represent the visual trajectory $(x, y)$ coordinates, while red dots indicate real-time robot coordinates upon receiving generated velocity commands. The figure also displays a sample of four images from the visual trajectory. }
    \label{linear}
\end{figure}

In this paragraph we describe tests made for pure translation. Throughout the robot movement, we gather the corresponding $(x, y)$ coordinates for each indexed image, representing them as blue points, as illustrated in Figure~\ref{linear}.

The collected visual trajectory is given to the model, the robot can follow the desired trajectory with minimal photometric error over a series of iterations, as depicted by the red dots.

We can observe a slight deviation in the robot path starting at the location marked by coordinates $(2.25, 2.510)$. These slight variations are permissible as long as the error, expressed as $\lvert I_{t} - I_{i} \rvert$, does not exceed the specified limit of $e_{m}$. If the main focus is to achieve higher accuracy, the value of $e_{m}$ can be reduced to minimize these deviations.

\subsection{Pure rotation}
\begin{figure}[!h]
    \centering
    \includegraphics[width=0.8\linewidth]{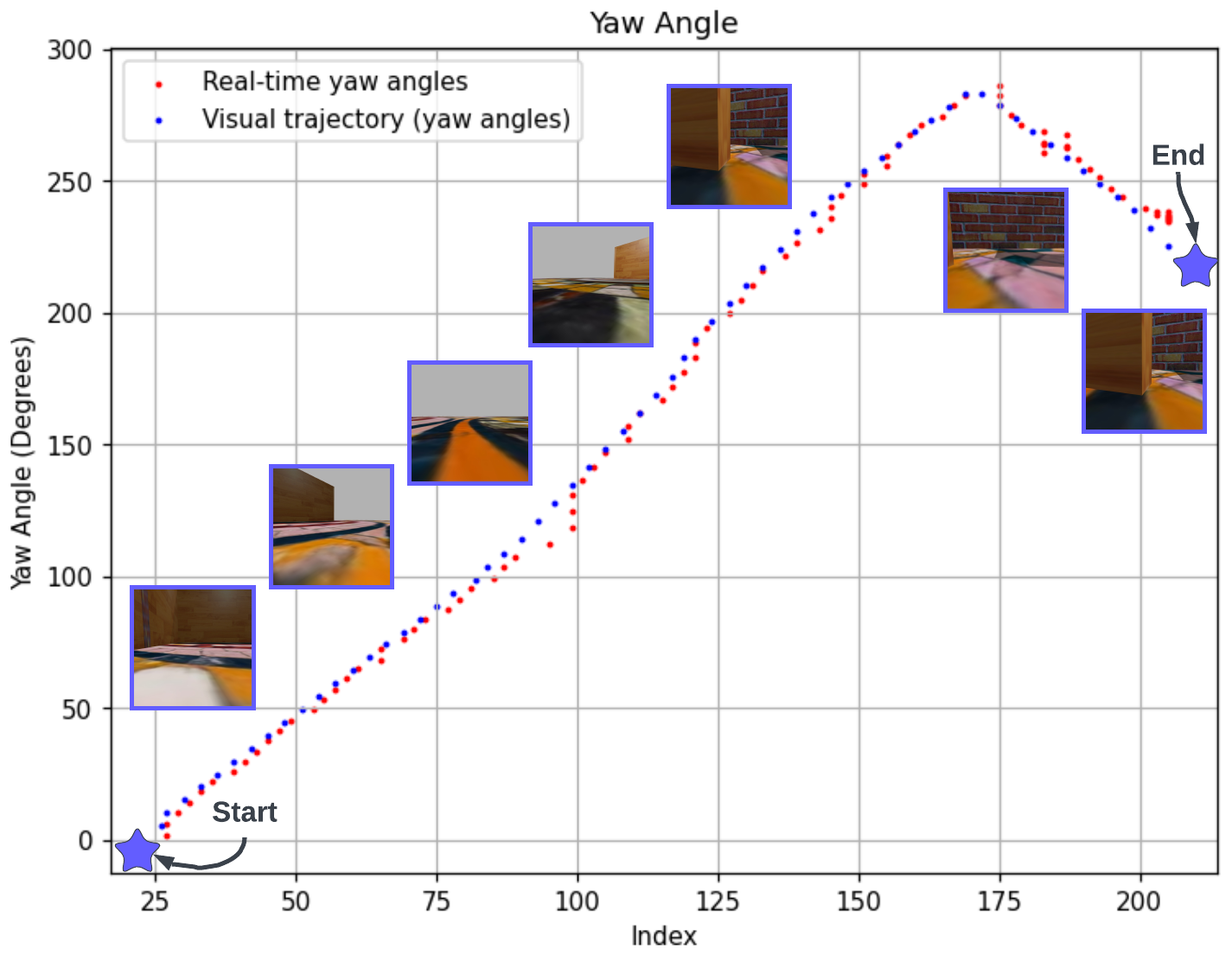}\caption{Evaluation of VelocityNet in Pure Rotation Scenario, blue points represent yaw angles gathered during visual trajectory tracking, while red dots illustrate real-time robot yaw angles in response to the predicted velocity commands,  with the index denoting subgoals indices. }
    \label{rotation}
\end{figure}

We collected a visual trajectory composed of images during the robot pure rotational movement, as represented in Figure~\ref{rotation} the robot did successfully reach each subgoal. The figure represents the robot yaw angles collected respectively for each subgoal index.

The collected visual trajectory represents the pure rotation of the robot in both directions. The robot starts rotating clockwise from subgoal 1 to subgoal 171 and then counterclockwise for the rest of the visual trajectory.
A similar deviation to what occurred during linear translation was observed within the subgoals range of 79 to 100, slight variations are acceptable as long as the error between the real time image and subgoal image, remains within the specified limit of $e_{m}$.

\subsection{Combined translation \& rotation}
\begin{figure}[!h]
    \centering
    \includegraphics[width=0.8\linewidth]{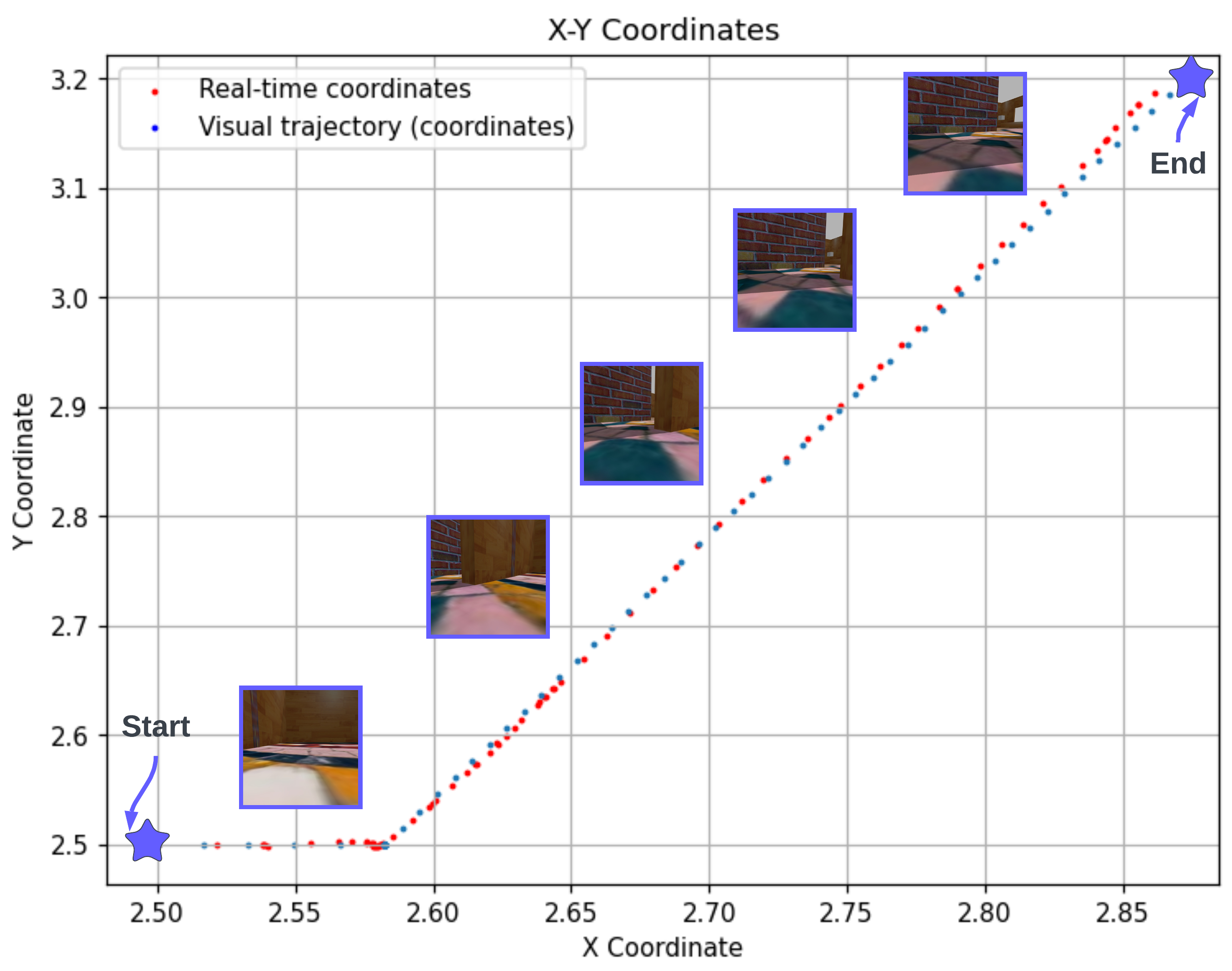}\caption{Evaluation of our predictive model VelocityNet in a (Translation + Rotation) scenario, blue points represent the visual trajectory $(x, y)$ coordinates, while red dots indicate real-time robot coordinates upon receiving generated velocity commands. The figure also displays a sample of images from the visual trajectory. }
    \label{rotrans}
\end{figure}

We gave our model a visual path that incorporates both linear translation and rotation. As shown in Figure~\ref{rotrans}, the robot initiates its motion by moving in a linear movement until it reaches the coordinates (2.57, 2.5). Afterward, it starts a rotational movement, and finally, it continues with linear translation until it reaches the end of the trajectory.

Our robot did successfully follow the desired path as shown with red dots in the same figure, however, as we approached the end of the trajectory the robot started to slip away from the desired subgoals.

\subsection{Statistical Analysis}

For each point $(x_i, y_i)$ in the visual trajectory coordinates, we find the nearest point $(x_{t}, y_{t})$ in the real-time coordinates and compute the error between the two points as the following:
\begin{equation}
\text{error}(x_i, y_i) = \sqrt{(x_i - x_{t})^2 + (y_i - y_{t})^2}
\label{error(x,y)}
\end{equation}

We use the same method for the pure rotation scenario; for each angle $\psi_i$ in the visual trajectory yaw angles, we find the nearest angle $\psi_t$ in the Real-time yaw angles, the error can be defined as:
\begin{equation}
\text{error}(\psi_i) = | \psi_i - \psi_t |
\label{error yaw}
\end{equation}
where i is the index of the subgoal image from the visual trajectory.\\

\begin{figure}[h!]
  \centering
  \begin{subfigure}{0.49\textwidth} % Adjust the width as needed
    \includegraphics[width=\linewidth]{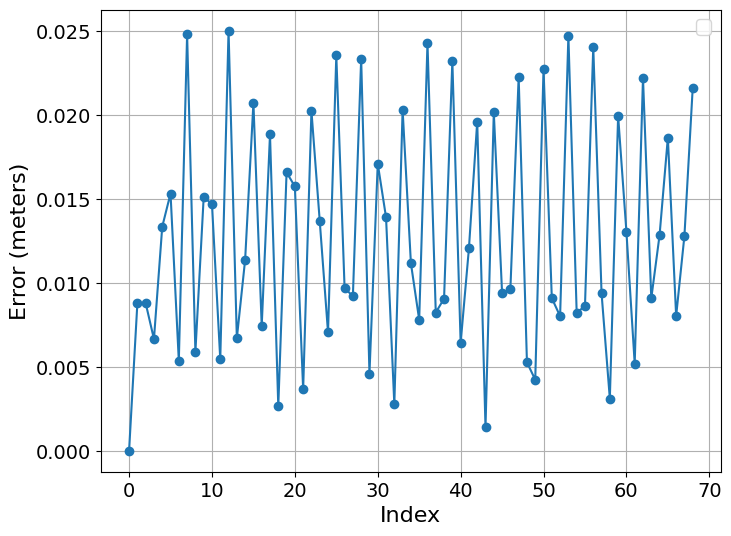}
    \caption{}
    \label{fig:subfig1}
  \end{subfigure}
 \begin{subfigure}{0.49\textwidth} % Adjust the width as needed
    \includegraphics[width=\linewidth]{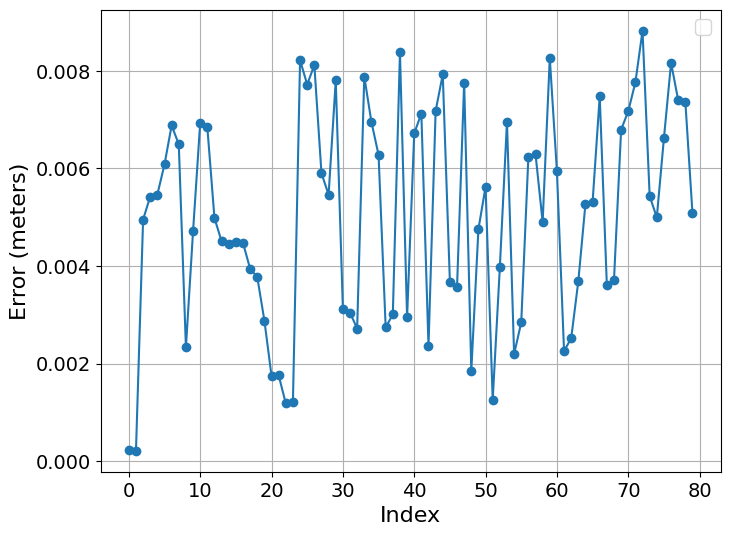}
    \caption{}
    \label{fig:subfig2}
  \end{subfigure}
  \begin{subfigure}{0.5\textwidth} % Adjust the width as needed
    \includegraphics[width=\linewidth]{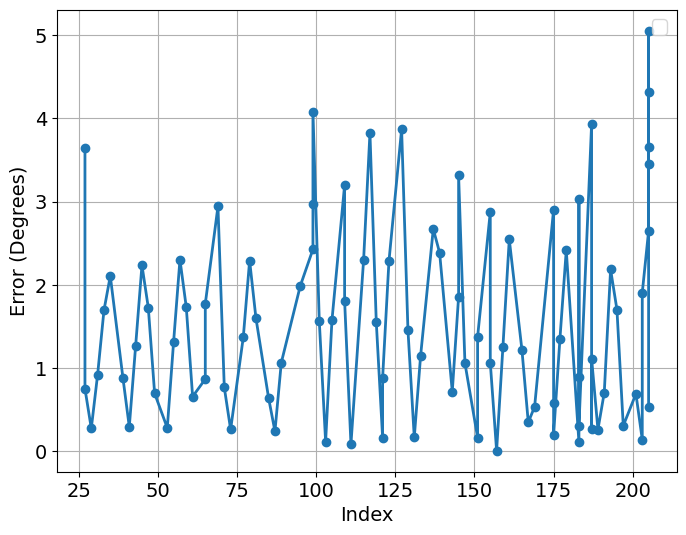}
    \caption{}
    \label{fig:subfig3}
  \end{subfigure}
\caption{Quantitative Evaluation of the Three Scenarios: (a) Pure Translation, (b) Translation + Rotation, and (c) Pure Rotational Motion, illustrating Error Plots for the selected subgoals.}
  \label{combined_figures}
\end{figure}
In Figure~\ref{combined_figures} (a) and (b), we can observe the plotted metric errors calculated using Equation~\ref{error(x,y)} for the selected subgoal indexes. Notably, in the pure translation scenario, these errors exhibit oscillations around the mean value calculated in Table~\ref{stattrans+rot}. These results demonstrate stability, which is an encouraging sign for a robust model that does not diverge. In the combined rotation and translation scenario, the errors exhibit overall stability, with a slight uptick in error values at the end of the trajectory.

In Figure~\ref{combined_figures} (c), we can see the yaw angle errors computed using Equation~\ref{error yaw} for the selected subgoal indexes. Similar to the pure translation scenario, the model exhibits indications of stability.

Tables~\ref{stattrans+rot} and~\ref{statrot} summarize the statistical data of the error results presented above. The mean errors across all scenarios are relatively low.

It's noteworthy that the standard deviation in the scenario involving pure rotation is significantly higher at 1.187 compared to the lower standard deviations of $6.98 \times 10^{-3}$ in the pure translation motion scenario and $2.203 \times 10^{-3}$ in the translation+rotation motion scenario. This difference in standard deviations suggests that in the rotation motion scenario, the robot's behavior tends to be less consistent, making its predictions less reliable, and the results more prone to variability.

\begin{table}[!h]
\centering
\caption{Error statistics for pure translation and translation + rotation scenarios}
\begin{tabular}{|l|c|c|c|c}
\hline
& Pure translation & Translation + Rotation  \\
\hline
Min (meters) & $3.37 \times 10^{-6}$  &  $2.13 \times 10^{-4}$ \\
\hline
Max (meters) & 0.025  & $8.825 \times 10^{-3}$  \\
\hline
MSE (meters) & 0.012 & $5.04 \times 10^{-3}$ \\ 
\hline
Standard deviation (meters) & $6.98 \times 10^{-3}$  & $2.203 \times 10^{-3}$ \\ 
\hline
\end{tabular}
\label{stattrans+rot}
\end{table}
\begin{table}[!h]
\centering
\caption{Error statistics for pure rotation scenario}
\begin{tabular}{|l|c|c|c|c}
\hline
& Rotation \\
\hline
Min (degrees) & $6.36 \times 10^{-3}$ \\
\hline
Max (degrees) & 5.053   \\
\hline
MAE (degrees) & 1.578 \\ 
\hline
Standard deviation (degrees) & 1.187   \\ 
\hline
\end{tabular}
\label{statrot}
\end{table}

%%%%%%%%%%%%%%%%%%%%%%%%%%%%%%%%%%%%%%%%%%%%%%%%%%%%
\section{CONCLUSION}
\label{sec:conclusion}
%%%%%%%%%%%%%%%%%%%%%%%%%%%%%%%%%%%%%%%%%%%%%%%%%%%%
% TODO
We proposed a novel way to use MPC policies with
deep neural networks, and applied them to visual navigation using an RGB camera. VelocityNet is a neural network that is offline trained with the same goals as an MPC controller. This allows VelocityNet to learn how to generate velocities that minimize the disparity between the current image of the robot and the target images along a visual trajectory. The offline training scheme allows also the use of less computational power. Our experiments showed that a visual navigation system based on VelocityNet can robustly follow visual trajectories in simulation.

Achieving a reasonable balance between how accurately our robot follows its path and the time it takes to complete the trajectory can be a bit challenging. This is because there are instances where the robot may deviate from the path at certain points. Therefore, identifying the appropriate value for $e_m$ is a complex task and could be a topic for future research.
VelocityNet is a novel deep-learning approach for training robots to navigate using visual cues. It is more efficient and robust than traditional methods and can be used in a wider range of environments. 

One of the key features of VelocityNet is that it can be trained offline. This has several advantages, including the ability to navigate in environments where online training is not feasible due to time constraints. Additionally, VelocityNet can be trained on images that are collected from the environment, which are less time-consuming to collect.

We can further improve our model by incorporating various enhancements, such as the integration of obstacle avoidance. GoNet~\cite{8594031} for instance can be used by monitoring the horizon for obstacle detection and predicting whether the robot-predicted velocities may result in a collision.
The metric errors obtained in the simulation show promising results. In the future, we plan to implement our experiments in real-world scenarios. We will also plan to investigate more advanced methods for novel view synthesis. Methods such as Neural Radiance Fields (NeRF) could be a promising tool regarding this issue. We will also increase the prediction horizon and compare our method to online approaches.

% perspectives: 
% results shown in simulation will be tested on a real robot. 
% obstacle detection is another perspective. This can be done using GONet\cite{}

\bibliographystyle{splncs04}
\bibliography{mybibliography}

\end{document}